# ArterialNet: Reconstructing Arterial Blood Pressure Waveform with Wearable Pulsatile Signals, a Cohort-Aware Approach

Sicong Huang, *Student Member*, *IEEE*, Roozbeh Jafari, *Fellow*, *IEEE*, and Bobak J. Mortazavi, *Senior Member, IEEE*

***Abstract—Goal:*** **Continuous arterial blood pressure (ABP) waveform is invasive but essential for hemodynamic monitoring. Current non-invasive techniques reconstruct ABP waveforms with pulsatile signals but derived inaccurate systolic and diastolic blood pressure (SBP/DBP) and were sensitive to individual variability.** ***Methods:*** **ArterialNet integrates generalized pulsatile-to-ABP signal translation and personalized feature extraction using hybrid loss functions and regularizations.** ***Results:*** **ArterialNet achieved a root mean square error (RMSE) of 5.41 ± 1.35 mmHg on MIMIC-III, achieving 58% lower standard deviation than existing signal translation techniques. ArterialNet also reconstructed ABP with RMSE of 7.99 ± 1.91 mmHg in remote health scenario.** ***Conclusion:*** **ArterialNet achieved superior performance in ABP reconstruction and SBP/DBP estimations with significantly reduced subject variance, demonstrating its potential in remote health settings. We also ablated ArterialNet's architecture to investigate contributions of each component and evaluated ArterialNet's translational impact and robustness by conducting a series of ablations on data quality and availability.**

***Impact Statement*—ArterialNet generates continuous ABP waveforms that derive accurate SBP/DBP points through wearable pulsatile waveforms including PPG and bioimpedance in both ICU and remote health settings.**

## I. Introduction

ARTERIAL blood pressure (ABP) waveforms are continuous, pulsatile representations of blood pressure fluctuations in the artery. These biosignals provide key determinants of cardiac functions such as stroke volume and cardiac output; they also serve as early indicators of potential major adverse cardiovascular (CVD) events such as cardiogenic shock. Consequently, timely ABP waveforms are particularly valuable in critical perioperative and postoperative settings [4].

However, obtaining ABP waveform requires invasive A-line catheterization, risking bleeding and infection [6]. Despite its values in forecasting CVD risks [2], its invasive nature limits its use primarily to to intensive care unit (ICU) settings. Although cuff-based solutions measure systolic and diastolic blood pressure (SBP/DBP) points, they cannot comprehensively monitor cardiac functions [6]; while volume-clamping techniques like Finapres offers noninvasive cardiac monitoring but are limited by sleep disturbance, motion artifacts, and discomfort during nocturnal, physical activity, and long-term monitoring [7].

Therefore, non-invasive methods for continuous, invasive ABP waveforms monitoring are needed. This work develops a system that uses non-invasively captured signals to estimate A-line ABP waveforms to enable broader clinical applications.

### *A. Related Works*

Recent non-invasive ABP monitoring methods use deep learning (DL)-based sequence-to-sequence (seq2seq) techniques to transform peripheral pulsatile signals to synthetic ABP waveforms [6]. Due to limited ABP waveform accessibility, most existing approaches are developed and evaluated on physiological signals from ICU patients. The Medical Information Mart for Intensive Care III (MIMIC-III) waveform dataset is the most frequently utilized resources [6, 8]. Hill et al. developed a U-Net framework to reconstruct ABP waveforms using both electrocardiography (ECG) and photoplethysmography (PPG) to capture temporal dynamics, reporting cohort-level findings [2]. Ibtehaz et al. also used a U-Net framework to reconstruct ABP waveforms from single-channel PPG recordings, reporting cohort-level findings and analysis [3]. While average group outcomes validated ABP reconstruction feasibilitys, they failed to capture individual variabilities. Such personalization is necessary for accurate modeling and a focus of this work.

Conversely, individual-specific performance in non-invasive BP monitoring varies significantly: effective for some but generalizes poorly, obscuring whether covariates reflect underlying cardiovascular dynamics [9].

This work was supported in part by the National Institutes of Health, under grant 1R01HL151240-01A1 *(Corresponding Author: Sicong Huang)*.

Sicong Huang and Bobak J. Mortazavi are with the Department of Computer Science and Engineering, Texas A&M University, College Station, TX 77840 USA (e-mail: [siconghuang, bobakm]@tamu.edu).

Roozbeh Jafari is with Lincoln Laboratory, Massachusetts Institute of Technology, Lexington, MA, USA, Laboratory for Information & Decision Systems (LIDS), Massachusetts Institute of Technology, Cambridge, MA, USA, Department of Electrical and Computer Engineering, Texas A&M University, College Station, TX, USA, and School of Engineering Medicine, Texas A&M University, Houston, TX, USA (e-mail: rjafari@mit.edu, roozbeh.jafari@ll.mit.edu)

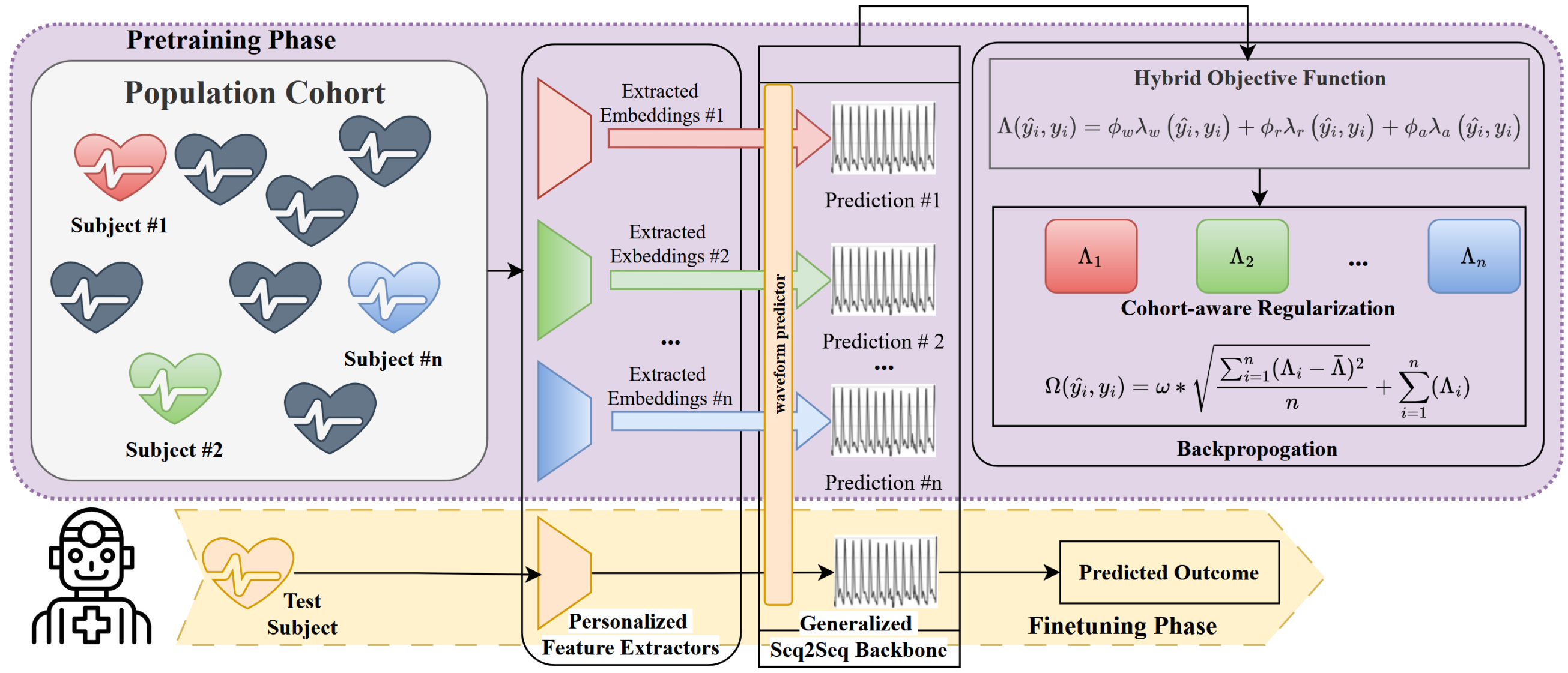

Figure 1: Overview of ArterialNet, a multi-layered pulsatile to ABP framework pretrained with a population cohort and then calibrated to a new individual. In ArterialNet, all components but seq2seq backbone are aware of who the data/embeddings came from during backpropagation. L denotes the hybrid objective function, l denotes each sub loss function, $\hat{y}$ denotes predicted waveform, $y$ reference waveforms, ϕ and ω denote weight hyperparameters.

Moreover, no previous work has evaluated framework transferability across different input modalities or architectures.

### B. Contribution

We introduce ArterialNet, a flexible framework that pretrains a seq2seq backbone on population data and fine-tunes per individual for ABP reconstruction. ArterialNet scales with input modalities, supporting both single-channel and pulse transit time (PTT) schemes while learning shared latent representations. ArterialNet introduces a hybrid loss function optimizing continuous ABP reconstruction, key fiducial point estimation, incorporating waveform correlation and alignment, and employing cohort-aware regularizations to minimize individual variations for improved ABP waveforms and derived SBP/DBP values.

We evaluated the performance of ArterialNet on two pulsatile modalities from both ICU and non-ICU datasets, conducted ablation studies to validate each component's importance, and assessed the robustness via data augmentation and masking experiments representing real-world scenarios.

We presented the preliminary version at the IEEE-EMBS International Conference on Biomedical and Health Informatics [10], and now extends this work in several ways: 1) expanding the feature extractor to compute PTT features and morphologies [11], 2) expanding the hybrid objective combining waveform reconstruction, correlation, and alignment losses, 3) conducting ablation studies on ArterialNet components, and 4) evaluating ArterialNet's robustness under various augmentations.

## II. Materials and Methods

This section details ArterialNet's two-stage paradigm: pretrainingfrom multi-person vitals to maximize generation capability, fintuning for a unseen person minimal data, as illustrated in Figure 1SBP/DBP[II].

### A. Personalized Feature Extractor

The personalized feature extractor is an adaptive interface that converts biosignals of variable frequencies and conditions into standardized embeddings to the seq2seq backbone.

ABP waveform reflects both immediate and historical cardiovascular dynamics [12]. Therefore, the feature extractor uses dilated causal convolutions to learn complex patterns from the multi-beat pulsatile input sequences in both single-sequence and PTT fashion.

### B. Seq2Seq Backbone

Standardized feature extraction allows ArterialNet to incorporate virtually any supervised backbones. We selected two backbones: U-Net and transformer due to their state-of-the-art performance across several related works [2, 3, 5, 13, 14]. While we trained seq2seq backbones from scratch, they can be pretrained to extend impact.

### C. Hybrid Objective Function

Let predicted waveforms be $\hat{y}$, reference waveform be $y$, and root-mean-squared error (RMSE) be reconstruction criterion θ, and waveform reconstruction loss be $\theta(\hat{y}, y)$.

We also computed 5 statistical features of the waveform as penalty objective (Ψ) [mean, standard deviation, skewness, minimum, and maximum] from y and $\hat{\Psi}$ from $\hat{y}$, respectively. With a weight hyperparameter α [0,1] to adjust penalty importance, our updated waveform objective function ($\lambda_w$) became:

$$\lambda_w(\hat{y}, y) = (1 - \alpha) * \theta(\hat{y}, y) + \alpha * \theta(\hat{\Psi}, \Psi). \quad (1)$$

We further conditioned the losses on waveform correlation by measuring waveform correlation loss ($\lambda_r$) between $\hat{\theta}$ and θ

[II] The Pytorch implementation of ArterialNet is available at https://github.com/stmilab/ArterialNet/tree/arterialnet+.

as Pearson's correlation and transforming the correlation coefficients into the range [0, ∞) where 0 denoted perfectly correlatedwith a custom reciprocal function. To prevent division by zero, we introduced a small constant c. The loss became:

$$\lambda_r(\hat{y}, y) = \frac{1}{2\times\left(\frac{\sum(\widehat{y_i}-\bar{\hat{y}})(y_i-\bar{y})}{\sqrt{\sum(\hat{y}-\bar{\hat{y}})^2\sum(y-\bar{y})^2}}+1\right)+c} \quad (2)$$

Since cardiac waveforms are time-series, we assessed the alignment quality by calculating the soft minimum over all alignment losses using soft-dynamic-time-warping (soft-DTW) loss ($\lambda_a$) [15], which defines a positive smoothing parameter $\gamma$, to calculate the alignment matrix $\beta$, as outlined:

$$\lambda_a(\hat{y}, y) = -\gamma \log \sum e^{-\beta/\gamma}, \beta \in \mathrm{dtw}(\hat{y}, y). \quad (3)$$

The hybrid objective function $\Lambda$ integrated them to provide a holistic measurement of reconstruction quality. We applied weight hyperparameters $[\phi_w, \phi_r, \phi_a]$ to balance the weights of computed losses. We found [1, 10, 0.01] to provide stable gradient descents experimentally.

$$\Lambda(\hat{y}, y) = \phi_w\lambda_w(\hat{y}, y) + \phi_r\lambda_r(\hat{y}, y) + \phi_a\lambda_a(\hat{y}, y). \quad (4)$$

### D. Cohort-aware Regularization

Training a cohort of $n$ individuals requires minimizing the aggregated hybrid loss $\Lambda(\hat{y}_i, y_i)$ (as $\Lambda_i$ for simplicity) across all $i$.

We added personalized regularizations based on inter-individual variances to promote subject-independent features. Using Krueger et al.'s variance risk extrapolation (REx) theory, we calculated the regularized loss ($\Omega$) of all training losses [$\Lambda_1$, $\Lambda_2$, ..., $\Lambda_n$] by adding weighted variance ($\sigma^2$) to the sum ($\Sigma$) in equation (5) [16]:

$$\Omega(\hat{y}_i, y_i) = \omega * \sqrt{\frac{\sum_{i=1}^{n}(\Lambda_i-\bar{\Lambda})^2}{n}} + \sum_{i=1}^{n}(\Lambda_i). \quad (5)$$

We defined the weight hyperparameter $\omega$ of range [0, ∞) to control the strength of regularization.

## III. Results

### A. Evaluation Metrics

We assessed both ABP reconstruction quality and SBP/DBP estimations by performing cardiac segmentation and fiducial point extraction to identify predicted and reference SBP/DBP points from generated and reference ABP waveforms, respectively [17]. Then, we evaluate waveform performance using average values ($\mu$) and standard deviation ($\sigma$) of root-mean-square-error (RMSE), mean-absolute-error (MAE), and Pearson's correlation coefficient (R) between reconstructed and reference waveforms. We then evaluated SBP/DBP in $\mu$ and $\sigma$ using the same metrics [RMSE, MAE, R].

We evaluated ArterialNet against several related studies with different seq2seq backbones: long short-term memory (LSTM) [1], V-Net [2], U-Net [3], and Transformer [5][III].

### B. Data Collection

#### 1) ICU Dataset Collection

We selected the MIMIC-III waveform dataset for our ICU setting evaluation. The dataset contained 22,317 pulsatile PPG, ECG, and associated ABP waveform records [18]. We excluded patients with extreme hemodynamics (e.g., extreme respiratory rate, oxygen saturation, etc.), invalid recordings, or narcotic or illicit drug use, organ failure, or major internal bleeding during check-in.

Consequently, we constructed a cohort of 61 patients (34 females), median age of 65 (range 25-87) and shared patient list on GitHub.

#### 2) Non-ICU Dataset Collection

We recruited 20 healthy participants (ages 18–40; 9 males, 11 females; IRB2020-0090F, Texas A&M). Each wore a bio-

Table 1: Performance evaluation of proposed ArterialNet versus related studies on both ABP waveform reconstruction and physiological estimations.

| Table of ABP Reconstruction Evaluation on ICU Data Collections collected with PPG and/or ECG | | | | | | | | | | | | | |
|---|---|---|---|---|---|---|---|---|---|---|---|---|---|
| Method | Dataset | Pulsatile Modality | # of Subjects | Total data (hours) | Performance Metrics (RMSE and MAE in mmHg, no unit for R) | | | | | | | | |
| | | | | | ABP (SD) | | | SBP (SD) | | | DBP (SD) | | |
| | | | | | RMSE | MAE | R | RMSE | MAE | R | RMSE | MAE | R |
| LSTM [1] | MIMIC | PPG | 42 | - | 6.04 (3.26) | 5.98 | 0.95 | 2.58 | - | - | 1.98 | - | - |
| V-Net [2] | MIMIC III | PPG + ECG | 264 | 1923+ | 5.82 | - | 0.96 | - | - | - | - | - | - |
| U-Net [3] | MIMIC II | PPG | 942 | 354 | - | 4.60 (5.04) | - | - | 5.73 (9.16) | - | - | 3.45 (6.15) | - |
| Transformer [5] | MIMIC | PPG | 241 | 150~241 | - | - | - | - | 4.01 (5.93) | 0.90 | - | 2.97 (3.87) | 0.84 |
| ArterialNet U-Net | MIMIC III | PPG | 56 | 733 | *5.78 (1.45)* | *4.52 (1.91)* | *0.92 (0.04)* | *5.76 (1.93)* | *4.30 (1.97)* | *0.89 (0.05)* | *4.65 (1.68)* | *3.38 (1.68)* | *0.87 (0.04)* |
| | MIMIC III | PPG + ECG | 56 | 733 | *5.47 (1.94)* | *4.16 (2.17)* | *0.91 (0.03)* | *5.99 (2.18)* | *4.39 (1.98)* | *0.88 (0.03)* | *5.38 (1.55)* | *4.52 (1.36)* | *0.83 (0.03)* |
| ArterialNet Transformer | MIMIC III | PPG | 56 | 733 | ***5.41 (1.35)*** | ***4.17 (1.29)*** | *0.91* ***(0.04)*** | *5.26* ***(1.35)*** | *4.15* ***(1.32)*** | ***0.90 (0.03)*** | *4.01* ***(1.55)*** | *3.17* ***(1.37)*** | ***0.88 (0.01)*** |
| | MIMIC III | PPG + ECG | 56 | 733 | *5.73 (1.76)* | *4.18 (1.19)* | *0.89 (0.03)* | *6.73 (2.44)* | *5.18 (2.03)* | *0.86 (0.03)* | *4.54 (2.35)* | *4.58 (1.47)* | *0.88 (0.03)* |

[III] Due to each baseline study being conducted on a separate and undisclosed cohort or dataset, we adopted their individual findings, alongside our reported results.

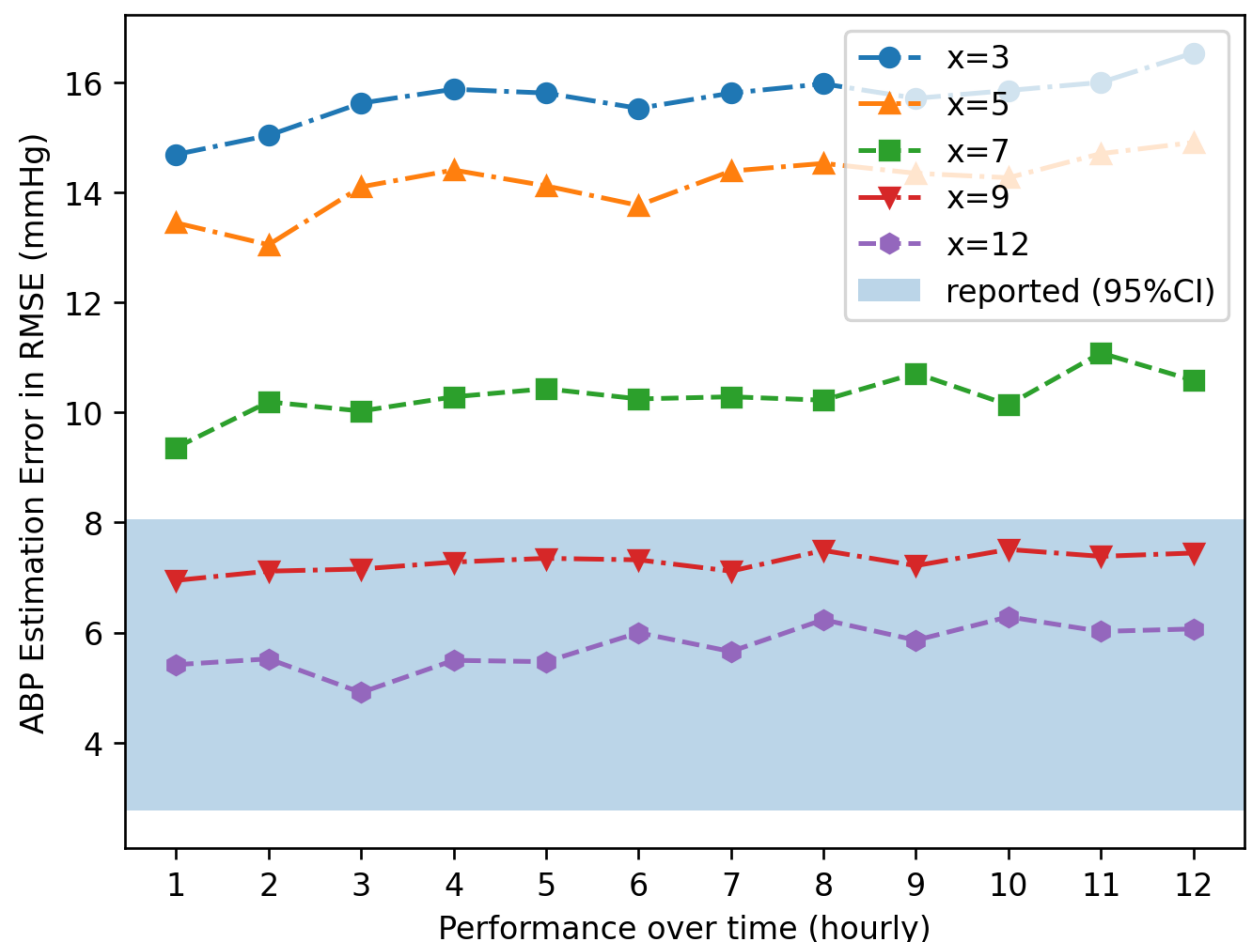

Figure 2: Visualization of ArterialNet's performance over time with (x) hours of calibration.

impedance (Bio-Z) wearable to collect peripheral pulsatile PPG and Bio-Z signals coupled with Finapres NOVA for continuous ABP waveform [19]. Everyone completed 8-minute protocols consisting of 0.5 minutes of rest, 3 minutes of hand-gripping to raise BP, 1 minute of placing a foot in ice water to keep BP elevated after hand-gripping, and 3.5 minutes of rest to recover BP. Each participant repeated the protocols 4 times per visit and had 7 visits scheduled at least 24 hours apart.

### *C. Data and Model Preparation*

We preprocessed both datasets with Pulse2AI and reported the details of preprocessing in Supplementary[11].

Our two-stage training paradigm started with supervised pretraining. We pretrained ArterialNet with data from five holdout MIMIC-III patients [27172, 47874, 94897, 56038, 82574] four for training and one for validation. The choice of five was selected to reflect the realities of data scarcity often encountered in real-world applications. By using a small, representative subset, we simulated data-scarce conditions to ensure that robustness and generalizability. Pretrainingis detailed in Supplementary.

## IV. Discussion

### *A. Performance Evaluation*

#### *1) MIMIC-III ICU Experiment*

We finetuned the pretrained ArterialNet on each of the remaining 56 patients using 80-20 individual-level split and applied the same hyperparameter tuning as pretraining. We reported ArterialNet's performance on both waveform and SBP/DBP estimations versus baselines on Table 1.

Employing our ArterialNet with both backbones demonstrated overperformance with significantly lower SD on derived SBP/DBP estimations. Furthermore, ArterialNet demonstrated superior generalizability as our study cohort contained more subjects and longer sequences. We further evaluated the quality of ArterialNet's derived SBP/DBP via Bland-Altman Analysis in Supplementary.

#### *2) Bio-Z Non-ICU Experiment*

We finetuned the ArterialNet on each participant's first five trials and evaluated the rest, performed the same hyperparameter tuning, and reported the results in Table 2. We measured the calibration per individual and reported the mean and standard deviation of these evaluation metrics.

We demonstrated ArterialNet could generate correlated ABP waveforms and derived SBP/DBP and proved the feasibility of ABP waveform reconstruction via peripheral Bio-Z pulsatile. To the best of our knowledge, this is the first ABP reconstruction study with bio-impedance pulsatile on healthy participants.

### *B. Robustness Evaluation*

We conducted several ablations on data quality and availability to assess ArterialNet's robustness, encompassing both data splitting level and waveform level data augmentations, to represent various real-world scenarios.

#### *1) Ablating Data Split*

Standard train-validation split may fail in real-world deployment due to various issues: unrealistic calibration time, sustaining performance over time, implausible temporal precedence, and inability to adjust to unknown BP ranges.

**Reducing calibration:** Calibration time denoted the number of sequential train data needed for finetuning prior to deployment to satisfy the chronological order. We first conducted sequential data splits to evaluate ArterialNet's performance, progressively reducing the training data size on both datasets and reported them in Table 3. We confirmed that performance increased as training size expanded, hence our model correctly learned more information from more data.

We also evaluated ArterialNet's on limited calibration time by incrementally testing its performance from 30 minutes to 12 hours of calibration before deployment in Figure 2. Using 95% confidence interval (CI) of reported performance as target interval, we found ArterialNet needed 12 hours of calibration to achieve its reported performance.

**Sustaining performance over time**: We further evaluate the deployment stability per hour for the next 12 hours to investigate the changes in model quality/degradation over time and reported performance over time on [3, 5, 7, 9, 12] hours of calibration in Figure 3.

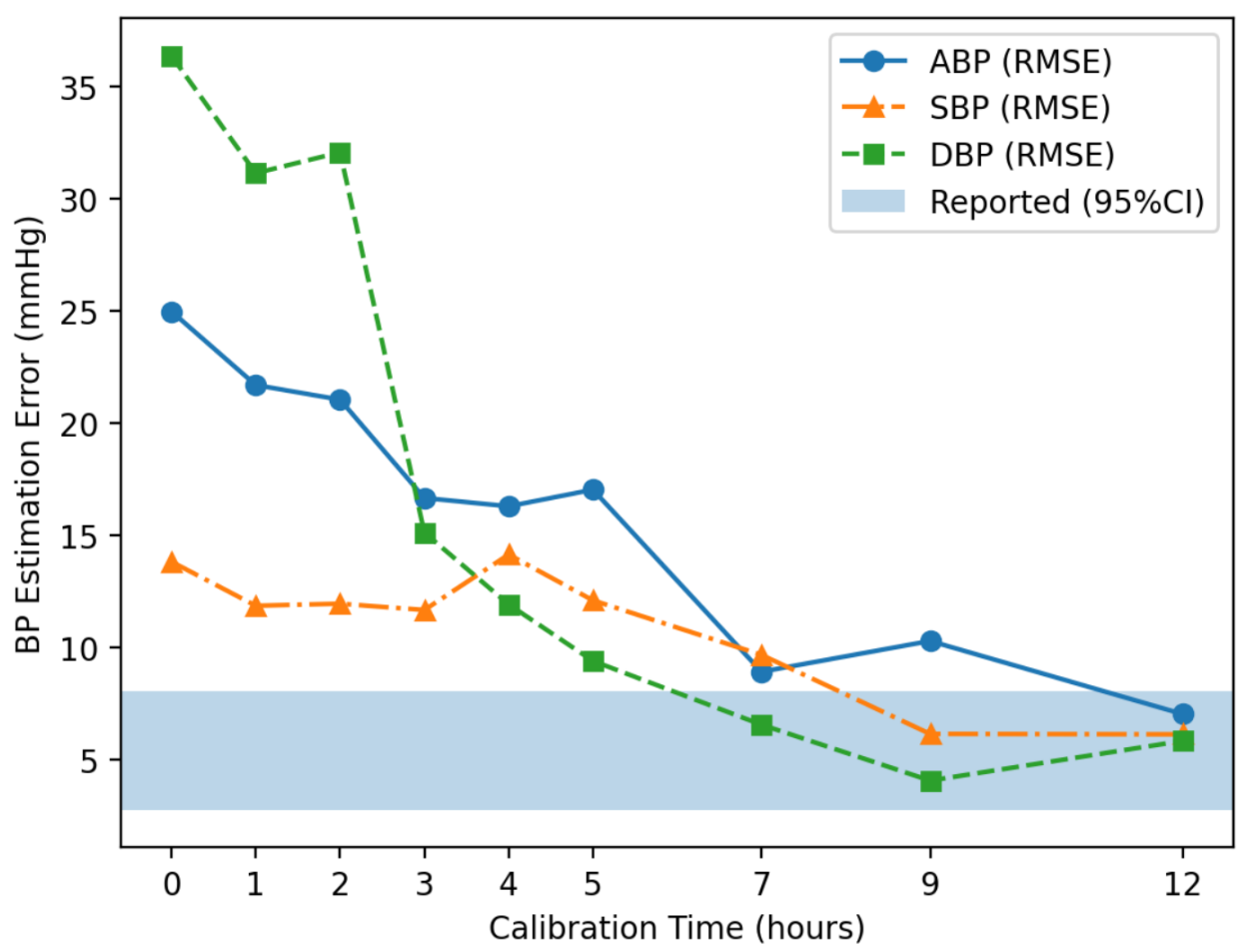

Figure 3: Ablation study of ArterialNet robustness on different calibration (training) time on MIMIC dataset

Table 2: Performance evaluation of proposed ArterialNet on non-ICU data collections using Bio-Z pulsatile signal.

| Method | Pulsatile modality | # of Subjects | Total data (hours) | Performance Metrics (RMSE and MAE in mmHg, no unit for Pearson's R) | | | | | | | | |
|---|---|---|---|---|---|---|---|---|---|---|---|---|
| | | | | ABP (SD) | | | SBP (SD) | | | DBP (SD) | | |
| | | | | RMSE | MAE | R | RMSE | MAE | R | RMSE | MAE | R |
| ArterialNet U-Net | Single Bio-Z | 20 | 54 | *8.66 (1.13)* | *6.90 (1.00)* | *0.44 (0.10)* | *12.27 (1.66)* | *10.57 (1.21)* | *0.51 (0.10)* | *8.03 (1.78)* | *6.63 (1.34)* | *0.39 (0.09)* |
| | Dual Bio-Z | 20 | 54 | *8.17 (1.42)* | *6.56 (1.23)* | *0.49 (0.13)* | *13.19 (1.91)* | *11.10 (1.41)* | *0.51 (0.11)* | *7.93 (1.91)* | *6.16 (1.44)* | *0.34 (0.14)* |
| ArterialNet Transformer | Single Bio-Z | 20 | 54 | *8.91 (1.14)* | *6.93 (0.81)* | *0.43 (0.12)* | *14.10 (1.81)* | *10.69 (1.54)* | *0.49 (0.10)* | *9.08 (1.91)* | *7.85 (1.60)* | *0.36 (0.10)* |
| | Dual Bio-Z | 20 | 54 | ***7.79*** *(1.91)* | ***6.01*** *(1.09)* | ***0.49*** *(0.21)* | ***11.94*** *(2.19)* | ***10.44*** *(1.99)* | ***0.52*** *(0.14)* | *7.77 (1.94)* | ***6.03*** *(1.45)* | *0.37 (0.15)* |

**Masking BP ranges:** We tested ArterialNet's generalizability by masking specific BP ranges during training and evaluating its performance on masked data. SBP ranges are masked in 10-mmHg intervals (90-100, 100-110, 110-120, 120-130, 130-140) and results are reported in Table 4. On MIMIC, ArterialNet remained stable across masking ranges of 100~120 mmHg, a healthier and more common range but dropped in sparse ranges, with SBP/DBP estimations declining rapidly. Similar trends appeared on Bio-Z, too.

*2) Augmenting Waveform*

Neural networks coud overfit, memorizing data patterns and producing susceptible outputs with noisy inputs. This phenomenon poses significant concern, given reliability is paramount in medical modeling. We assessed ArterialNet's robustness via several custom augmentation.

**Embedding noise:** We added Gaussian noise to each waveform (same length, mean, half SD) with a multiplier controlling amplitude (0 = original). Results in Table 5 demonstrated ArterialNet remained robust with lower noise rate (0.1). When but under heavy noise, ArterialNet collapses, that ArterialNet constructed a mapping for signal translation, instead of merely memorizing waveform patterns through proof by contradiction.

Table 3: Ablation study of ArterialNet robustness on unknown BP

| Masking SBP ranges (RMSE in mmHg, no unit for Pearson's R) | | | | | | |
|---|---|---|---|---|---|---|
| MIMIC | ABP (SD) | | SBP (SD) | | DBP (SD) | |
| | RMSE | R | RMSE | R | RMSE | R |
| 90-100 | 9.63 (4.17) | 0.87 (0.03) | 13.83 (3.77) | 0.41 (0.02) | 5.80 (3.19) | 0.41 (0.02) |
| 100-110 | **7.75 (1.99)** | **0.92 (0.02)** | **8.36 (2.29)** | **0.71 (0.02)** | **5.69 (2.33)** | **0.64 (0.02)** |
| 110-120 | **7.56 (2.19)** | **0.91 (0.02)** | **8.15 (3.01)** | **0.69 (0.02)** | **6.63 (2.49)** | **0.62 (0.01)** |
| 120-130 | 11.67 (2.32) | 0.86 (0.03) | 13.02 (2.24) | 0.58 (0.02) | 7.91 (2.95) | 0.33 (0.02) |
| 130-140 | 11.15 (3.41) | 0.87 (0.03) | 13.19 (3.11) | 0.47 (0.02) | 7.85 (2.93) | 0.31 (0.02) |
| Bio-Z | | | | | | |
| 100-110 | 35.30 (8.99) | 0.36 (0.33) | 24.95 (2.00) | 0.28 (0.34) | 19.75 (4.92) | 0.37 (0.06) |
| 110-120 | 21.03 (7.16) | 0.42 (0.20) | 18.46 (3.55) | 0.33 (0.14) | 22.22 (6.13) | 0.47 (0.16) |
| 120-130 | 18.12 (1.52) | 0.31 (0.08) | 14.57 (0.83) | 0.14 (0.09) | 11.04 (1.73) | 0.18 (0.07) |
| 130-140 | 16.69 (0.86) | 0.44 (0.17) | 12.25 (1.56) | 0.26 (0.18) | 14.29 (2.37) | 0.42 (0.16) |

**Masking within a cardiac cycle:** Since each cardiac cycle of pulsatile waveform started with a primary pulse wave and followed by a reflection wave, we ablated the ArterialNet by masking either the earlier half (pulse wave) or the latter half (reflection wave) of the pulsatile waveforms and reported the performance in Table 5. On MIMIC, ArterialNet used features from both segments, while on Bio-Z, masking reflection waves caused collapse—indicating reliance on reflection waves.

**Masking previous/current cardiac cycles:** Capturing information from a longer sequence of multiple, recurrent cardiac cycles of pulsatile data is one of ArterialNet's major contributions. We evaluated the role of recurrent cycles by masking previous or current cycles (Table 5). ArterialNet remained accurate without previous cycles, but masking current cycles weakened predictions—showing they are essential.

*3) Analyzing Neural Network Layers*

Understanding the contribution of individual components to overall improvement offers critical insights about each component's importance. We ran ablation studies on MIMIC to assess component contributions. Results showed major SBP/DBP gains from the hybrid objective and highlighted each loss's role. Cohort-aware regularization in our two-stage paradigm further improved performance. Extra features added little benefit, but consistent results confirmed the feature extractor's reliability. Full details are shared in Supplementary.

## V. CONCLUSION

ArterialNet is a multi-layer pulsatile to ABP framework pretrained on a population cohort and finetuned to individuals to improve ABP reconstruction and SBP/DBP estimation qualities. When evaluated on the in-clinic dataset, ArterialNet outperformed all baselines across three metrics by considerable margins. On the non-clinical dataset, ArterialNet also proved its ability to capture pulsatile information across different pulsatile signals and the potential for ABP monitoring for remote health settings.

In this extension, we extended the scope feature extractor, improved the efficacy of hybrid objective function with waveform reconstruction, correlation, and alignment losses, demonstrated the contributions of each component, and evaluated its translational impact and robustness via various data augmentations.

Table 4: Ablation study of ArterialNet robustness on data split

| Shrinking train-test split (RMSE in mmHg, no unit for Pearson's R) | | | | | | |
|---|---|---|---|---|---|---|
| MIMIC | ABP (SD) | | SBP (SD) | | DBP (SD) | |
| | RMSE | R | RMSE | R | RMSE | R |
| 10-90 | 11.72 (3.17) | 0.85 (0.02) | 14.17 (3.89) | 0.57 (0.02) | 8.72 (2.56) | 0.45 (0.05) |
| 30-70 | 13.46 (1.53) | 0.85 (0.03) | 17.46 (1.44) | 0.54 (0.11) | 6.87 (2.56) | 0.39 (0.03) |
| 50-50 | 8.92 (1.55) | 0.90 (0.02) | 9.67 (1.95) | 0.66 (0.02) | 6.56 (2.38) | 0.55 (0.06) |
| 70-30 | 6.07 (1.96) | 0.94 (0.01) | 6.15 (1.44) | 0.88 (0.01) | 4.07 (1.78) | 0.81 (0.07) |
| *80-20* | *5.41 (1.35)* | *0.91 (0.04)* | *5.26 (1.35)* | *0.90 (0.03)* | *4.01 (1.55)* | *0.88 (0.01)* |
| 90-10 | 4.97 (1.44) | 0.95 (0.01) | 4.15 (1.24) | 0.90 (0.01) | 4.07 (1.54) | 0.89 (0.02) |
| Bio-Z | | | | | | |
| 10-90 | 20.65 (4.74) | 0.38 (0.11) | 22.54 (2.94) | 0.33 (0.11) | 11.36 (1.10) | 0.02 (0.01) |
| 30-70 | 18.22 (2.31) | 0.463 (1.47) | 22.38 (4.13) | 0.43 (0.08) | 14.21 (4.50) | 0.05 (0.09) |
| 50-50 | 17.47 (1.92) | 0.54 (0.05) | 19.10 (1.88) | 0.49 (0.05) | 11.17 (2.39) | 0.08 (0.04) |
| 70-30 | 9.07 (2.94) | 0.47 (0.02) | 12.91 (1.71) | 0.50 (0.21) | 9.51 (3.46) | 0.18 (0.11) |
| *80-20* | *7.79 (1.91)* | *0.49 (0.21)* | *11.94 (2.19)* | *0.52 (0.14)* | *7.77 (1.94)* | *0.37 (0.15)* |
| 90-10 | 7.97 (1.97) | 0.67 (0.03) | 11.67 (1.39) | 0.69 (0.10) | 6.19 (4.30) | 0.52 (0.08) |

### *A. Limitations*

The method was validated on retrospective data, which may not capture real-world variability. Despite using ICU and non-ICU datasets, the findings may not generalize to broader populations or settings.

## ETHICS

This study was approved by the Texas A&M University IRB (IRB2020-0090F). All participants provided informed consent. MIMIC-III data used in this study is publicly available and fully de-identified, complying with HIPAA regulations.

## CONFLICT OF INTEREST

Dr. Roozbeh Jafari is a member of the editorial board for OJEMB; other authors declare no conflict of interest.

## AUTHOR CONTRIBUTION

S.H., R.J., and B.J.M. conceived the idea. S.H. processed the data, compiled the models, and conducted the experiments, while R.J. and B.J.M. supervised the entire process. The manuscript was written with contributions from all authors, and all authors have approved the final version of the manuscript.